\documentclass[a4paper]{article}

\usepackage{INTERSPEECH2020}
\usepackage{float}
\usepackage{makecell}
\usepackage{multirow}
\usepackage{graphicx}
\usepackage{caption}
\usepackage{bm}
\usepackage{enumitem}
\usepackage{amsmath}
\makeatletter
\def\maketag@@@#1{\hbox{\m@th\normalfont\normalsize#1}}
\makeatother

\title{Knowledge Distillation and Data Selection for Semi-Supervised Learning in CTC Acoustic Models}
\name{Prakhar Swarup, Debmalya Chakrabarty, Ashtosh Sapru, Hitesh Tulsiani, Harish Arsikere, \\ Sri Garimella}
\address{
Amazon Alexa, Bangalore, India}
\email{\{swarupps,debmac,sapru,hittul,arsikere,srigar\}@amazon.com}

\begin{document}

\maketitle
\begin{abstract}
Semi-supervised learning (SSL) is an active area of research which aims to utilize unlabelled data in order to improve the accuracy of speech recognition systems. The current study proposes a methodology for integration of two key ideas: 1) SSL using connectionist temporal classification (CTC) objective and teacher-student based learning 2) Designing effective data-selection mechanisms for leveraging unlabelled data to boost performance of student models. Our aim is to establish the importance of good criteria in selecting samples from a large pool of unlabelled data based on attributes like confidence measure, speaker and content variability. The question we try to answer is: Is it possible to design a data selection mechanism which reduces dependence on a large set of randomly selected unlabelled samples without compromising on Word Error Rate (WER)? We perform empirical investigations of different data selection methods to answer this question and quantify the effect of different sampling strategies. On a semi-supervised ASR setting with 40000 hours of carefully selected unlabelled data, our CTC-SSL approach gives 17\% relative WER improvement over a baseline CTC system trained with labelled data. It also achieves on-par performance with CTC-SSL system trained on order of magnitude larger unlabeled data based on random sampling.

\end{abstract}

\section{Introduction}

In recent years, introduction of Recurrent Neural Networks (RNNs), in particular, Long Short Term Memory (LSTM) RNNs \cite{Graves2005FramewisePC} have been shown to outperform feed-forward neural networks \cite{Sak2015LearningAF} and significantly improve the performance of automatic speech recognition (ASR) systems. More recently, it has been shown that LSTMs when trained with CTC loss followed by sequence discriminative training can achieve state-of-the-art performance on various large-scale acoustic modelling tasks \cite{Graves2006ConnectionistTC,Senior2015_ASRU}. However, a primary bottleneck of these frameworks is their dependence on availability of large amounts of labeled data. Since collection and transcription of large amounts of speech data is costly and time-consuming, techniques to leverage large amounts of unlabelled data for acoustic modeling are explored under the framework of semi-supervised learning (SSL)\cite{Manohar2018_ICASSP,huang2016semi-supervised}.

SSL for ASR involves automatic generation of labels for unlabelled data and leveraging them along with labelled data for building ASR systems. Self-training has been shown to be an effective SSL framework where an initial system is trained on labelled data to generate machine transcriptions for unlabelled data \cite{Huang2013SemisupervisedGA,Thomas2013_ICASSP,chen2020semisupervised,Shinoda2012_APSIPA,Quitry2016_SLT}. However, the effectiveness of such systems gets constrained by the ``goodness" of the initial system and ``quality" of machine transcripts. Teacher-Student based learning \cite{Hinton2015_NIPS} originally proposed in the context of knowledge distillation (KD) for model compression is another approach used for SSL. KD involves using a strong teacher model trained on labelled data to generate  labels which are used to train a student model to match teacher output distribution. 

The effectiveness of KD has been well established in speech recognition tasks \cite{Kim2018ImprovedTF,Chebotar2016DistillingKF,Watanabe2017StudentteacherNL,Fukuda2017EfficientKD}. However, conventional frame-level based KD approaches may not be directly feasible for CTC trained models due to `alignment-free' nature of CTC loss function. Modifications have been proposed in recent studies which either attempt to align teacher and student models' spike timings \cite{Kurata2019, Kurata2018_SLT} or propose KD at the level of sequences instead of frames \cite{Huang2018KnowledgeDF,Takashami2018_ICASSP}. The cost associated with aligning output spikes from teacher and student model make the frame-level based approaches computationally expensive when using large amounts of unlabelled data. Existing studies on sequence-level KD for CTC entail a certain degree of complexity in sampling n-best hypothesis \cite{Takashami2018_ICASSP} or computing forward-backward posteriors from teacher model \cite{Huang2018KnowledgeDF}. A simplified version of sequence-level KD has been studied in the context of neural machine translation (NMT) \cite{kim-rush-2016-sequence}, which essentially collapses the entire label sequence space to a single 1-best sequence. In this work, we simplify this assumption further by approximating the entire label sequence space by a single 1-best sequence obtained by concatenating frame-level argmax output from teacher model. Our results demonstrate that such a framework provides better WER as compared to a standard self-training based approach.

In a recent study in KD \cite{Hari2019_ICASSP}, it was shown that augmenting labelled data with 1 million hours of randomly selected unlabelled data improves relative WER by $10$-$20$\%. However, processing such large quantities of unlabelled data and subsequent training requires access to powerful compute and storage resources and leads to significant increase in training time. The prime motivation for this work is to seek answer to the following question: Instead of large amounts of randomly selected unlabelled data, is it possible to design an intelligent data selection scheme which can achieve similar WER gains but with lesser data?

Majority of data selection approaches studied in the past have relied on confidence scores, entropy based confidence measure and local-global entropy minimization among others for bootstrapping initial seed model with additional unlabelled data \cite{Chen2019,YU2010433} in the self-learning framework. To the best of our knowledge, data selection in the context of teacher-student learning has not been explored extensively.  We present an empirical study of the role of different data selection mechanisms based on confidence and Natural Language Understanding (NLU) based domain distribution as well as speaker and content diversity on WER of CTC-SSL based student model. Overall, our CTC-SSL framework achieves 17\% relative WER improvement over a baseline CTC system with labelled data only and on-par performance with a CTC-SSL model trained on an order of magnitude higher randomly sampled unlabelled data.


\section{SSL via Knowledge Distillation for CTC}
\label{sec:ctc_kd_seq}

One of the tasks of speech recognition involves mapping a sequence of frame-level labels (called `path' and denoted as $\bm{\pi}$) into a label sequence (denoted as $\bm{h}$) of length equal to or less than the number of frames. In the CTC framework \cite{Graves2006ConnectionistTC}, a path is converted into a label sequence by introducing deletion of repeated as well as blank labels. This conversion is referred to as `CTC mapping' with function $B$, where $\bm{h}=B(\bm{\pi})$. Since multiple possible paths can be mapped into an identical label sequence, conditional probability of label sequence $\bm{h}$ given input sequence $\bm{x}$ is defined by Equation (\ref{eq:ctc_basic}).

\begin{equation}
\footnotesize
    P(\bm{h}|\bm{X}) = \sum_{\bm{\pi}\in B^{-1}(\bm{h})}P(\bm{\pi}|\bm{X}) = \sum_{\bm{\pi} \in B^{-1}(\bm{h})} \prod_{t=1}^{N_{X}} P(\pi_{t} | \bm{x}_{t})
    \label{eq:ctc_basic}
\end{equation}


Posterior probability $P(\pi_{t}|\bm{x}_{t})$ of label $\pi_{t}$ at frame $t$ is typically modeled with a RNN. $N_{X}$ denotes the number of frames in utterance $\bm{X}$. In this case, training objective is to minimize negative log-likelihood of label sequence given input frames as given by $L_{CTC} = -ln(P(\bm{h}|\bm{X}))$.\\



For unlabelled dataset, our aim is to generate pseudo-labels from a strong teacher model. Teacher-Student learning based knowledge distillation (KD) has been studied for CTC acoustic models primarily from a model compression standpoint \cite{Takashami2018_ICASSP}. The objective of KD as shown in Equation (\ref{eq:frame_level_kd}) is to train a smaller student model using the output of a stronger teacher model as training labels.
 
\begin{equation}
     \footnotesize
    L_{CTC-KD_{frame}} = -\sum_{t=1}^{N_{X}}\sum_{\pi_{t} \in Z}P_{T}(\pi_{t}|\bm{x}_{t})ln(P_{S}(\pi_{t}|\bm{x}_{t}))
    \label{eq:frame_level_kd}
\end{equation}

Where $P_{T}(\pi_{t}|\bm{x}_{t})$, $P_{S}(\pi_{t}|\bm{x}_{t})$ denote the teacher and student model output respectively and $Z$ denotes the set of all labels. This approach is referred to as frame-level KD. In sequence-level based KD a student model is trained by minimizing cross entropy loss between probability distributions of label sequences $\bm{h}$ generated by teacher and student model as shown in Equation (\ref{eq:ctc_kd_seq}).

\begin{equation}
  L_{CTC-KD_{seq}} = -\sum_{\bm{h} \in H}P_{T}(\bm{h}|\bm{X})ln(P_{S}(\bm{h}|\bm{X}))
    \label{eq:ctc_kd_seq}
\end{equation}

Where $H$ denotes the set of all possible label sequences. Substituting Equation (\ref{eq:ctc_basic}) into Equation (\ref{eq:ctc_kd_seq}) for $P_{T}(\bm{h}|\bm{X})$, we see that computation of $L_{CTC-KD_{seq}}$ involves a summation over all paths $\bm{\pi}$ such that $B(\bm{\pi}) \in H$ which is a computationally expensive operation. Hence, similar to \cite{kim-rush-2016-sequence}, we approximate the summation by its maximum value i.e. a single 1-best path to get Equation (\ref{eq:max_approx}).


\begin{equation}
	\footnotesize
    L_{CTC-KD_{seq}}  \approx - \max_{\bm{\pi}} \prod_{t=1}^{N_{X}} P_{T}(\pi_{t}|\bm{x}_{t}) ln(P_{S}(B(\bm{\pi})|\bm{X}))
    \label{eq:max_approx}
\end{equation}

For further simplification, we apply the greedy decoding approximation suggested by Equation (4) in \cite{Graves2006ConnectionistTC} which approximates the best path by a path formed by concatenating the most probable label at each time step. Due to the spiky nature of CTC posteriors, $P_{T}(\pi_{t}|\bm{x}_{t})$ can be approximated by a Kronecker delta function at $\arg \max P_{T}(\pi_{t}|\bm{x}_{t})$. Hence, substituting $\bm{\pi}^{*}$ as path formed by concatenating the most probable labels $\pi_{t}^{*}$ at each time frame $t$ from teacher model's output, Equation (\ref{eq:max_approx}) can be transformed into Equation (\ref{eq:ctc_kd_seq_final}).


\begin{equation}
    L_{CTC-KD_{seq}} \approx - ln(P_{S}(B(\bm{\pi}^{*})|\bm{X}))
    \label{eq:ctc_kd_seq_final}
\end{equation}

Overall, this approximation to sequence-level knowledge distillation for unlabelled data can be described in following steps: 
\begin{enumerate}
    \item Sample frame level labels $\pi_{t}^{*}$ for unlabelled data by taking an $\text{argmax}$ over teacher posterior vector at each time step $t$.
    \item Convert frame-level label sequence $\bm{\pi}^{*}$ to CTC labels by applying CTC mapping operator $B(\bm{\pi}^{*}) = \bm{h}^{*}$
    \item Use this label sequence $\bm{h}^{*}$ for conventional CTC model training.
\end{enumerate}

Combining loss functions for labelled data $D_{L}$ and unlabelled data $D_{U}$, we get the overall training loss function as shown in Equation (\ref{eq:train_loss}).
\begin{equation}
    L = -\sum_{(\bm{X},\bm{h}) \in D_{L}} ln(P_{S}(\bm{h}|\bm{X}))-\sum_{\bm{X} \in D_{U}} ln(P_{S}(\bm{h}^{*}|\bm{X}))
    \label{eq:train_loss}
\end{equation}

\section{Data Selection}
\label{sec:data_selection}

We discuss data selection strategies used in our SSL experiments below:

\subsection{Utterance-level Confidence Scores}
Utterance-level confidence scores, as described in our earlier work \cite{Swarup2019}, are used to select data for SSL experiments (details will be discussed in the next section). Utterance-level confidence scores are produced by a logistic regression classifier, which takes hand-crafted decoder features of an utterance as input and produces a posterior probability estimate that indicates whether ASR hypothesis corresponding to utterance is correct (WER=0) or not (WER$\neq$0). Input features for training the confidence model contain ASR decoder information such as utterance posterior, acoustic cost, language model cost, number of words, maximum depth of n-best output, frame length and statistics on the number of arcs and nodes explored in decoding trellis. The confidence model is trained using the cross-entropy criterion. 

\subsection{Utterance-level Domain Information}

Utterance-level domain information, estimated by our internal Natural Language Understanding (NLU) system, is also used to select data for SSL experiments in this study (details will be discussed in the next section). This is because ASR accuracies vary across domains (such as Music, Weather and Information etc.) due to domains having varying complexity of utterances. We demonstrate the effect of selecting domain-specific SSL data on per-domain ASR accuracies.

\subsection{Common filters}
\label{sub_sec:other_filters}
Apart from utterance-level confidence and domain information, additional filters are also employed to optimize the amount of unlabelled data:
\setlist{nolistsep}
\begin{itemize}[noitemsep]
    \item \textbf{No Wakewords}: We do not consider wakeword-only (“alexa”) utterances for SSL data selection as they are redundant and do not add content diversity.
    \item \textbf{Max samples per Content}: We sample a maximum of 50 utterances having identical 1-best recognition result.
    \item \textbf{Max samples per Device}: To increase the number of devices and diversity, we select a maximum of 50 utterances per device. This ensures that frequently used devices do not dominate overall unlabelled data. 
\end{itemize}
    
\section{Experiments and Results}

Experimental setup used in this study is described in Table \ref{tab:exp_setup}. AM training consists of cross-entropy training, which produces a seed AM for the final stage CTC training. The training data for building baseline AMs consist of anonymized labelled dialect-specific English data from Alexa’s production traffic. In case of SSL based AMs, additional unlabelled data is interleaved with the labelled data used for building baseline AM. In addition, the teacher AM training leverages multi-dialect CE and CTC training phases, where 45000 hours of labelled data is pooled across multiple English dialects, prior to dialect-specific sMBR training to boost its accuracy. For decoding, we use dialect-specific Language Models (LMs) that are trained on both production and external data sources.

\begin{table}[h]
\resizebox{\columnwidth}{!}{%
\begin{tabular}{cl}
\hline
\textbf{Training Framework} & Tensorflow 1.13 \\ \hline
\textbf{Feature representation} & \begin{tabular}[c]{@{}l@{}}3 * 256 dimensional \cite{Sak2015_arxiv}\\ Short-Time Fourier Transform\end{tabular} \\ \hline
\textbf{Label representation} & \begin{tabular}[c]{@{}l@{}}2608 Senones \\ (Hybrid CTC-HMM) \cite{Senior2015_ASRU}\end{tabular} \\ \hline
\textbf{Student training strategy} & Cross entropy -\textgreater CTC \\ \hline
\textbf{Student architecture} & \begin{tabular}[c]{@{}l@{}}FLSTM \cite{sainath2016_flstm}\\ Frequency LSTM : \\ Bidirectional, \\ Window = 48, Hop = 15, \\ Layers = 2, Units = 16. \\ Time LSTM: \\ Unidirectional, \\ Layers = 5, Units = 768\end{tabular} \\ \hline
\textbf{Dialects} & \begin{tabular}[c]{@{}l@{}}British English (en-GB)\\ Indian English (en-IN)\end{tabular} \\ \hline
\textbf{Student training corpus} & \begin{tabular}[c]{@{}l@{}}en-GB = 16000 hours\\ en-IN = 8000 hours\end{tabular} \\ \hline
\textbf{Evaluation samples} & \begin{tabular}[c]{@{}l@{}} en-GB = 32000 utterances \\ en-IN = 27000 utterances\end{tabular}  \\ \hline
\textbf{Teacher training strategy} & \begin{tabular}[c]{@{}l@{}}Cross entropy -\textgreater CTC \\ -\textgreater sMBR (Dialect data)\end{tabular} \\ \hline
\textbf{Teacher architecture} & \begin{tabular}[c]{@{}l@{}}FLSTM \cite{sainath2016_flstm}\\ Frequency LSTM : \\ Bidirectional, \\ Window = 48, Hop = 15, \\ Layers = 2, Units = 16\\ Time LSTM: \\ Bidirectional, \\ Layers = 5, Units = 1024\end{tabular} \\ \hline
\textbf{Teacher training corpus} & \begin{tabular}[c]{@{}l@{}} 45000 hours \\ en-US = 22500 hours \\ en-GB = 11000 hours \\ en-IN = 6500 hours \\ en-AU = 5000 hours \end{tabular}  \\ \hline
\end{tabular}%
}
\caption{CTC-SSL experimental setup}
\label{tab:exp_setup}
\end{table}

\subsection{CTC-SSL with random sampling}

The results of British (en-GB) and Indian (en-IN) English dialect-specific baseline and teacher models are shown in Table \ref{tab:250k_results}. It confirms that the teacher is indeed better than the baseline, and is a good candidate for CTC-SSL experiments. As a next step, we randomly select 250000 hours of data (similar to \cite{Hari2019_ICASSP}), produce pseudo-labels using appropriate teacher model, and build corresponding student model using CTC-SSL technique discussed in Section \ref{sec:ctc_kd_seq}. The results indicate that leveraging 250K hours of randomly selected unlabelled data and 8K-16K hours of labelled data using CTC-SSL framework yields 17\% WERR compared to a baseline that is trained on labelled data alone. This result also serves as an additional data point to evaluate the efficiency of data selection methods described in Section \ref{sec:data_selection}.   
\begin{table}[H]
    \centering
    \resizebox{\columnwidth}{!}{%
\begin{tabular}{|c|c|c|}
\hline
Model&en-GB WERR (\%)&en-IN WERR(\%)\\ \hline
Baseline&0.0&0.0\\
Teacher&27.8&28.6\\
250k randomly sampled SSL&17.2&17.6\\  
\hline
\end{tabular}
}
    \caption{WERR comparison between student, teacher and randomly sampled SSL model}
    \label{tab:250k_results}
\end{table}

We also compared our proposed Teacher-Student based approach with Self-training framework for en-IN. For Self-training, we built a multi-dialect unidirectional LSTM AM (similar to student’s architecture) using 45000 hours of labelled data which is used to machine-label 250000 hours of SSL data. As shown in Table. 3, Teacher-Student based CTC-SSL outperforms Self-training by yielding 6.3\% additional WERR. Therefore, we use BiLSTM based teacher AM for rest of SSL experiments in this paper. 

\begin{table}[H]
    \centering
\begin{tabular}{|c|c|}
\hline
Model&WERR(\%)\\ \hline
Baseline&0.0\\
Self-training&11.3\\
Teacher-Student&17.6\\ 
\hline
\end{tabular}
 \caption{WERR comparison between Teacher-Student and Self-training based SSL models for en-IN}
    \label{tab:teacher_student}
\end{table}

\subsection{Utterance confidence based sampling}
\begin{figure}[h]
	\centering
	\includegraphics[width=8cm]{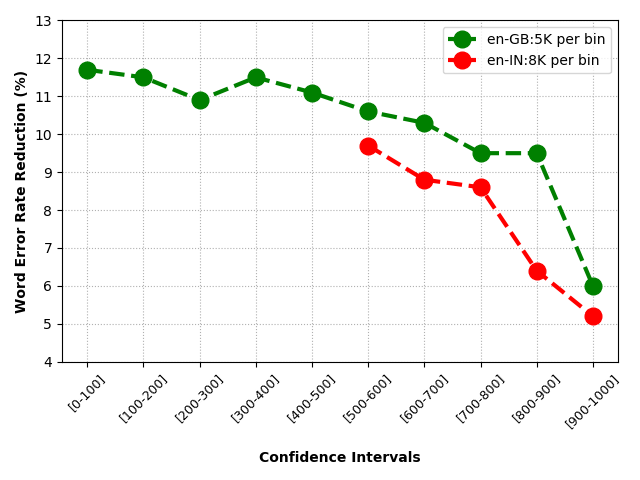}
	\caption{WERR comparison across student models after adding unlabelled data from utterance confidence bins}
	\label{tab:utt_conf_bin_results}
\end{figure}

In this section, we provide an empirical analysis of selecting unlabelled data based on confidence distribution for CTC-SSL training. The entire unlabelled data is divided into 10 uniformly spaced bins within [0-1000] ([0-1] scaled by 1000) based on confidence values extracted from anonymized production traffic.We sample 5000 hours of unlabelled data within each bin for en-GB. However, due to anonymized data sparsity constraints, 8000 hours of unlabelled data are sampled within [500-1000] for en-IN such that the total SSL data distribution remains similar across both the locales ($\sim$80\% less than randomly sampled 250k). Apart from utterance confidence, common filters described in Section \ref{sub_sec:other_filters} are also applied for data selection within each bin. We build CTC-SSL models for each of confidence bins following procedure discussed earlier. Figure \ref{tab:utt_conf_bin_results}  quantifies the effect of confidence based unlabelled data selection from each bin on the performance of CTC-SSL system (in terms of WERR) compared to baseline trained on labelled data only.

Figure \ref{tab:utt_conf_bin_results} clearly suggests that adding unlabelled data from low to mid range confidence distribution to CTC training positively impacts the performance of ASR system compared to baseline. As the sampling of unlabelled data moves from low to high confidence, it is observed that the contribution of SSL data to student performance drops from 11.7\% to 6\% for en-GB and 9.7\% to 5.2\% for en-IN. With respect to the student model, samples drawn from high confidence bins can be considered as `easy'  and hence adding them to existing labelled data is not beneficial. On the other hand, mid to low confidence bins contain `difficult' samples which when added to student model training will enable better generalization and improved KD from the teacher.\\

\noindent Next, we attempt to come up with a strategy to effectively combine data across multiple confidence bins. The approaches explored are:
\begin{itemize}
    \item Natural distribution (ND): Sample according to production pipeline distribution
    \item Uniform distribution (UD): Sample equally from all bins.
    \item Weighted sampling (WS): Sample based on WERRs in Table \ref{tab:utt_conf_bin_results}
\end{itemize}
The results, including large scale random selection (discussed earlier), are listed in Table \ref{tab:the_table}. 40000 hours of unlabelled data combined using WS and UD methods yield better results than that of ND. Interestingly, these smart unlabelled data selection methods with order of magnitude less data yield results comparable to 250K hours of randomly sampled experiment which reinforces the importance of data selection for semi-supervised learning in ASR.
\begin{table}[h]
    \centering
    \resizebox{\columnwidth}{!}{%
\begin{tabular}{|c|c|c|}
\hline
Model&en-GB WERR (\%)&en-IN WERR(\%)\\ \hline
Baseline&0.0&0.0\\
SSL with 250k&17.2&17.6\\
SSL with 40k [ND] &15.1&14.1\\
SSL with 40k [UD] &17.1&\bf{17.3}\\
SSL with 40k [WS] &\bf{17.4}&16.3\\
\hline
\end{tabular}
}
    \caption{WERR comparison across CTC-SSL models with different confidence bin based combination methods}
    \label{tab:the_table}
\end{table}

\subsection{Utterance domain based sampling}

We present empirical analysis of the effect of domain based sampling of unlabelled data in boosting the performance of ASR. Based on NLU domain extracted from anonymized production traffic, we sample 10000 hours and 8000 hours from 5 individual domains [D1$\cdots$D5] in en-GB and en-IN respectively and build CTC-SSL model by augmenting labelled data with individual domain unlabelled data. Apart from using domain information, other filters as described in Section \ref{sub_sec:other_filters} are also applied while selecting data for these experiments. Table \ref{tab:domains} validates the hypothesis that individual domain sampling improves the performance of corresponding domain without significant degradation across other domains, as supported by a clear diagonal behavior in both en-GB and en-IN specific WERR matrices. We also observe some cross domain degradation e.g. [D5 model-D2 test] in en-GB and  [D5 model-D3 test] in en-IN. We speculate that boosting a particular domain-specific data via SSL may result in data imbalance in training set. To tackle this issue, data across all the selected domains was combined (8000 hours * 5) such that each domain gets enough representation during CTC-SSL training. This 40000 hours of domain data ingested into training provides the greatest WERRs across all domains for en-IN as can be observed from comparing last row in Table \ref{tab:domains} against the other rows. However, for en-GB the per-domain WERRs from combined domain data are lesser than those obtained from individual domain based sampling. This is because for en-GB, individual domain based experiments had a better representation per-domain (10000 hours) as opposed to the combined domain experiment (8000 hours).

\begin{table}[H]
 \resizebox{\columnwidth}{!}{%
\begin{tabular}{|c|c|c|c|c|c|c|c|c|c|c|}
\hline
\multirow{1}{*}{}{Model}&\multicolumn{5}{c|}{en-GB WERR (\%)}&\multicolumn{5}{c|}{en-IN WERR (\%) }\\ \cline{2-11}
&D1&D2&D3&D4&D5&D1&D2&D3&D4&D5\\ \hline
Baseline&0.0&0.0&0.0&0.0&0.0&0.0&0.0&0.0&0.0&0.0\\	
D1&\bf{16.8}&5.0&-1.5&4.5&10.6&\bf{12.1}&6.5&2.2&11.6&0.5\\
D2&5.5&\bf{12.2}&5.2&7.3&12.8&8.2&\bf{10.0}&3.7&7.7&6.6\\
D3&1.6&0.6&\bf{21.2}&5.7&7.1&5.6&4.7&\bf{12.5}&7.7&4.5\\
D4&6.8&8.0&14.7&\bf{11.1}&8.6&8.1&5.8&4.3&\bf{15.5}&2.0\\
D5&10.1&-1.7&-0.8&5.2&\bf{21.9}&7.2&1.7&-4.1&5.7&\bf{8.5}\\
\makecell { Combined \\ Domains }&12.4&10.3&20.9&7.7&14.7&16.4&12.9&15.7&22.1&13.28\\
\hline
\end{tabular}
}
\caption{WERR comparison across CTC-SSL models based on individual domain based sampling}
    \label{tab:domains}
\end{table}

\section{Conclusions}

This paper presents an empirical study on large-scale semi-supervised learning for CTC acoustic models where a strong offline teacher model is used to generate pseudo-labels for unlabelled data. The unlabelled data is selected based on confidence and domain distribution as well as speaker and content variability. Experimental results on two different dialects reinforce the efficacy of teacher generated pseudo labels and the importance of intelligent data selection methods. It is observed that domain-specific unlabelled data has a strong impact on corresponding WER with little cross-domain impact signifying the importance of such a sampling strategy in boosting the performance of low resource domains. Future work in this direction would be to devise a strategy to leverage both confidence as well as domain diversity in a combined data sampling strategy for SSL. Another important future direction will be to study the impact of word level decoding which incorporates both lexicon and strong language model in improving the quality of teacher generated pseudo-labels. 

\bibliographystyle{IEEEtran}

\bibliography{mybib}

\end{document}